\documentclass{article}

\usepackage{PRIMEarxiv}

\usepackage[utf8]{inputenc} 
\usepackage[T1]{fontenc}    
\usepackage{hyperref}       
\usepackage{url}            
\usepackage{booktabs}       
\usepackage{amsfonts}       
\usepackage{nicefrac}       
\usepackage{microtype}      
\usepackage{lipsum}
\usepackage{fancyhdr}       

\usepackage[utf8]{inputenc}
\usepackage[dvips]{graphicx}
\usepackage{float} 
\usepackage[dvipsnames]{xcolor}
\usepackage{tikz}
\usetikzlibrary{fit}
\usepackage{amssymb,amsmath,array}
\usepackage{algorithm}
\usepackage{booktabs}
\usepackage{multirow}
\usepackage{algorithmic}
\title{Integrating Latent Block Models into Variational Autoencoders for Disentangled Representation Learning
}
\author{
  Anasse Essalih \\
  VEDECOM Institute \\
  DAVID Lab, UVSQ, Paris-Saclay University \\
  Versailles, France \\
  \texttt{anasse.essalih@vedecom.fr} \\
  \And
  Reda Khoufache \\
  TSM Unit \\
  UVSQ, Paris-Saclay University \\
  Versailles, France \\
  \texttt{reda.khoufache@uvsq.fr} \\
  \And
  Laurent Durville \\
  VEDECOM Institute \\
  Versailles, France \\
  \texttt{laurent.durville@vedecom.fr} \\
  \AND
  Mohamed-Cherif Rahal \\
  VEDECOM Institute \\
  Versailles, France \\
  \texttt{mohamed-cherif.rahal@vedecom.fr} \\
  \And
  Mustapha Lebbah \\
  DAVID Lab \\
  UVSQ, Paris-Saclay University \\
  Versailles, France \\
  \texttt{mustapha.lebbah@uvsq.fr} \\
  \And
  Hanene Azzag \\
  LIPN, UMR CNRS 7030 \\
  Sorbonne Paris Nord University \\
  Villetaneuse, France \\
  \texttt{azzag@univ-paris13.fr} \\
  \AND
  Walid Kheriji \\
  VEDECOM Institute \\
  Versailles, France \\
  \texttt{walid.kheriji@vedecom.fr} \\
  \And
  Fouad Hadj Selem \\
  VEDECOM Institute \\
  Versailles, France \\
  \texttt{fouad.hadjselem@vedecom.fr} \\
}
\begin{document}
\maketitle

\begin{abstract}
We introduce LBM-VAE, a novel framework that integrates Latent Block Models (LBMs) into Variational Autoencoders (VAEs) to jointly cluster observations and latent dimensions.  Beyond its clustering capabilities, LBM-VAE provides a block-structured latent representation that naturally supports disentangled representation learning. In contrast to conventional approaches that associate each factor of variation with an individual latent dimension, our framework allows factors of variation to be encoded across groups of latent dimensions. Experimental results demonstrate the promising clustering performance of LBM-VAE, while achieving competitive results in disentangled representation learning. LBM-VAE highlights the potential of combining co-clustering and variational inference to learn structured and interpretable latent representations.
\end{abstract}


\section{Introduction}

Disentangled representation learning seeks to recover latent codes in which distinct factors of variation are encoded separately~\cite{10.1109/TPAMI.2024.3420937}. VAE-based methods such as $\beta$-VAE~\cite{Higgins2016betaVAELB}, FactorVAE~\cite{pmlr-v80-kim18b} and $\beta$-TCVAE~\cite{NEURIPS2018_1ee3dfcd}
promote the aggregate posterior towards a factorial distribution, assuming that each factor is encoded in a single latent dimension. However, this assumption may be overly restrictive, as certain factors, such as color, naturally exhibit a multi-dimensional structure.
%
%
Existing approaches, however, do not learn such block structures from data. In disentangled representation learning, subspace-based approaches such as ISA-VAE~\cite{stühmer2019isavae} partition latent coordinates into independent subspaces. However, both the number and dimensionality of these subspaces are specified a priori. In deep clustering, mixture-based priors such as VaDE~\cite{3172077.3172161} cluster the observations, with each component spanning the entire latent space. MFCVAE~\cite{NEURIPS2021_48cb136b} extends this paradigm by introducing multiple facets, each associated with its own latent variables and mixture prior. Nevertheless, the latent dimensions assigned to each facet must be specified in advance by the user, which is not straightforward in practice. In contrast, we let the model discover this structure automatically, by replacing the conventional Gaussian prior with a Latent Block Model (LBM)~\cite{doi:https://doi.org/10.1002/9781118649480.ch2}. The LBM performs a simultaneous co-clustering of the latent code matrix, clustering observations into row clusters and latent dimensions into column clusters, with a distinct Gaussian per block. Given only an upper bound $L$ on the number of blocks, the model learns which latent dimensions are grouped together and the size of each block, so that groups of dimensions capturing common factors of variation emerge directly from the data. Our contributions are summarised as follows: (1) we introduce LBM-VAE, a variational autoencoder whose prior jointly infers partitions of both observations and latent dimensions (co-clustering); (2) we derive coordinate-ascent updates for the variational approximations of both partition posteriors and develop a mini-batch training scheme that updates the column partition using memoized sufficient statistics; and (3) we evaluate the proposed approach through experiments on MNIST and Shapes3D, demonstrating its ability to learn structured latent representations.

\section{Background}
\label{sec:lbm}

Let $n,p,K,L$ be positive integers, $[n]:=\{1,\dots,n\}$, $\Delta^{K-1}$ the
probability simplex of $\mathbb{R}^K$, and $\mathrm{Cat}(\boldsymbol{\pi})$ the
categorical distribution on $[K]$ with probabilities
$\boldsymbol{\pi}\in\Delta^{K-1}$.
The LBM~\cite{doi:https://doi.org/10.1002/9781118649480.ch2} assumes that a data
matrix $Y=(y_{ij})\in\mathbb{R}^{n\times p}$ is generated as follows:
\begin{equation}
  v_i \sim \mathrm{Cat}(\boldsymbol{\pi}), \qquad
  w_j \sim \mathrm{Cat}(\boldsymbol{\rho}), \qquad
  \boldsymbol{\pi}\in\Delta^{K-1},\ \boldsymbol{\rho}\in\Delta^{L-1},
  \label{eq:lbm-cat}
\end{equation}
\begin{equation}
  y_{ij}\mid v_i=k,\,w_j=l \;\sim\; \mathcal{N}(\mu_{kl},\sigma^2_{kl}),
  \qquad \mu_{kl}\in\mathbb{R},\ \sigma^2_{kl}>0 .
  \label{eq:lbm-emission}
\end{equation}
Memberships $v_i$ and $w_j$ are drawn independently, and the entries of a block $(k,l)$ are i.i.d.\ given them. With $\boldsymbol{\eta}=(\boldsymbol{\pi},\boldsymbol{\rho},
(\mu_{kl},\sigma^2_{kl})_{k,l})$, the likelihood is
\begin{equation}
  p(Y;\boldsymbol{\eta})
  =\sum_{\mathbf{v}\in[K]^n}\sum_{\mathbf{w}\in[L]^p}
   \prod_{i}\pi_{v_i}\prod_{j}\rho_{w_j}
   \prod_{i,j}\mathcal{N}\!\left(y_{ij};\mu_{v_iw_j},\sigma^2_{v_iw_j}\right).
  \label{eq:lbm-lik}
\end{equation}
This sum has $K^nL^p$ terms and the posterior $p(V,W\mid Y)$
does not admit a factorized form, so exact EM is intractable. Variational
EM~\cite{doi:https://doi.org/10.1002/9781118649480.ch2} restricts the posterior
to the mean-field family $q(V,W)=\prod_i q(v_i)\prod_j q(w_j)$ and alternates
coordinate-ascent updates of the responsibilities $\alpha_{ik}:=q(v_i=k)$ and
$\beta_{jl}:=q(w_j=l)$ with a closed-form M-step.

\section{The LBM-VAE framework}
\label{sec:model}

\subsection{Generative Model}
For observations $X=(x_1,\dots,x_n)$ and latent codes
$Z\in\mathbb{R}^{n\times p}$, the generative model is
\begin{subequations}
\label{eq:gen}
\begin{align}
  v_i &\sim \mathrm{Cat}(\boldsymbol{\pi}),\quad
  w_j \sim \mathrm{Cat}(\boldsymbol{\rho}),
    && i\in[n],\ j\in[p], \\
  Z_{ij} \mid v_i = k,\, w_j = l
    &\sim \mathcal{N}\big(\mu_{kl}, \sigma^2_{kl}\big),
    && i\in[n],\ j\in[p], \\
  x_i \mid Z_{i\cdot}
    &\sim \mathcal{D}_\theta(Z_{i\cdot}),
    && i\in[n],
\end{align}
\end{subequations}
where $\mathcal{D}_\theta$ is a decoder distribution parametrised by a neural
network with weights $\theta$, with density $p_\theta(\cdot\mid Z_{i\cdot})$.

\subsection{Inference and Evidence Lower Bound}
We use $q(Z,V,W\mid X)=\prod_{i}q_\phi(Z_{i\cdot}\mid x_i)\,q(v_i\mid x_i)\prod_{j}q(w_j\mid X)$ (mean field),
with $q_\phi( Z_{i\cdot}\mid x_i)=\mathcal{N}(\mathbf{m}_i,\mathrm{diag}\,\mathbf{s}_i^2)$. The ELBO reads
\begin{equation}
\begin{split}
  \mathcal{L}
  ={}&\textstyle\sum_{i}\mathbb{E}_{q_\phi}\big[\log p_\theta(x_i\mid Z_{i\cdot})\big]
  -\textstyle\sum_{i,j}\mathbb{E}_{q_\phi}\Big[\sum_{k,l}\alpha_{ik}\beta_{jl}
     \log\frac{q_\phi(Z_{ij}\mid x_i)}{\mathcal{N}(Z_{ij};\mu_{kl},\sigma^2_{kl})}\Big]\\
  &-\textstyle\sum_{i}\mathrm{KL}(\boldsymbol{\alpha}_i\,\|\,\boldsymbol{\pi})
  -\textstyle\sum_{j}\mathrm{KL}(\boldsymbol{\beta}_j\,\|\,\boldsymbol{\rho}).
\end{split}
\label{eq:elbo}
\end{equation}
As in~\cite{doi:https://doi.org/10.1002/9781118649480.ch2}, coordinate-ascent variational inference (CAVI)~\cite{bishop2006pattern} yields explicit updates for the variational approximations of both label posteriors:
\begin{align}
  e_{ij}^{kl} &:= \mathbb{E}_{q_\phi}\big[\log\mathcal{N}(Z_{ij};\mu_{kl},\sigma^2_{kl})\big]
  = -\tfrac12\Big(\log(2\pi\sigma^2_{kl})+\tfrac{(m_{ij}-\mu_{kl})^2+s_{ij}^2}{\sigma^2_{kl}}\Big), \nonumber\\
  \alpha_{ik} &\propto \pi_k\exp\Big(\sum_{j,l}\beta_{jl}\,e_{ij}^{kl}\Big),
  \qquad
  \beta_{jl}\propto\rho_l\exp(\ell_{jl}),
  \quad \ell_{jl}:=\sum_{i,k}\alpha_{ik}\,e_{ij}^{kl}.
  \label{eq:cavi}
\end{align}
The row posterior $\boldsymbol{\alpha}_i$ only involves observation $i$, whereas
$\boldsymbol{\beta}_j$ depends on the whole column $(m_{ij},s_{ij}^2)_{i\in[n]}$
of encoder outputs.

\subsection{Training and Memoized Variational Inference}
The row responsibilities $\boldsymbol{\alpha}_i$ are local: their
update~\eqref{eq:cavi} only involves observation $i$ and is computed on the
current mini-batch $\mathcal{B}_m$. In contrast, $\boldsymbol{\beta}$ is a
global variational parameter: its update~\eqref{eq:cavi} depends on $\phi$
through $\ell_{jl}$, which aggregates all observations, and since each training
step changes $\phi$, an exact update would require re-encoding the whole
dataset after every step.
We therefore adapt memoized online variational
inference~\cite{NIPS2013_d490d7b4} to our case, which updates global parameters from cached
per-batch sufficient statistics whose sums summarise the full dataset. For each
mini-batch $\mathcal{B}_m$, we store
\begin{equation}
  N^{(m)}_k=\sum_{i\in\mathcal{B}_m}\alpha_{ik},\quad
  S^{(1,m)}_{jk}=\sum_{i\in\mathcal{B}_m}\alpha_{ik}m_{ij},\quad
  S^{(2,m)}_{jk}=\sum_{i\in\mathcal{B}_m}\alpha_{ik}\big(m_{ij}^2+s_{ij}^2\big),
  \label{eq:batch-stats}
\end{equation}

which are indexed by latent coordinate $j$ and row cluster $k$, rather than by block.
With the
totals $N_k$, $S^{(1)}_{jk}$, $S^{(2)}_{jk}$ summed over mini-batches,
\begin{equation}
  \ell_{jl}=-\tfrac12\textstyle\sum_{k}\big(N_k\log(2\pi\sigma^2_{kl})
  +(S^{(2)}_{jk}-2\mu_{kl}S^{(1)}_{jk}+N_k\mu_{kl}^2)/\sigma^2_{kl}\big),
  \label{eq:ell}
\end{equation}
After the gradient step on $\mathcal{B}_m$, its statistics are recomputed with the
updated encoder and swapped into the totals (old contribution subtracted, new one
added), and $\boldsymbol{\beta}$ is recomputed from~\eqref{eq:ell}; $\boldsymbol{\eta}$ is updated at the end of each epoch.
%


\section{Experiments}
\begin{figure}[H]
  \centering
  \includegraphics[width=\textwidth]{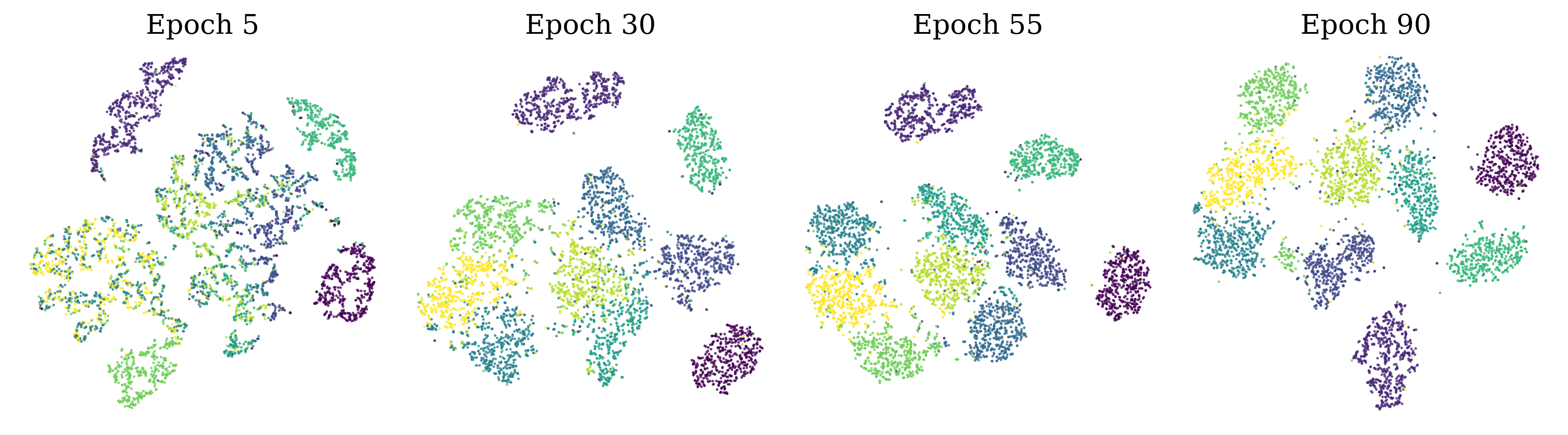}
  \caption{t-SNE projections of the MNIST test-set latent means at epochs
5, 30, 55 and 90, coloured by digit class (LBM prior activated at epoch 5).}
  \label{fig:tsne}
\end{figure}

\begin{figure}[!t]
  \centering
  \includegraphics[width=\textwidth]{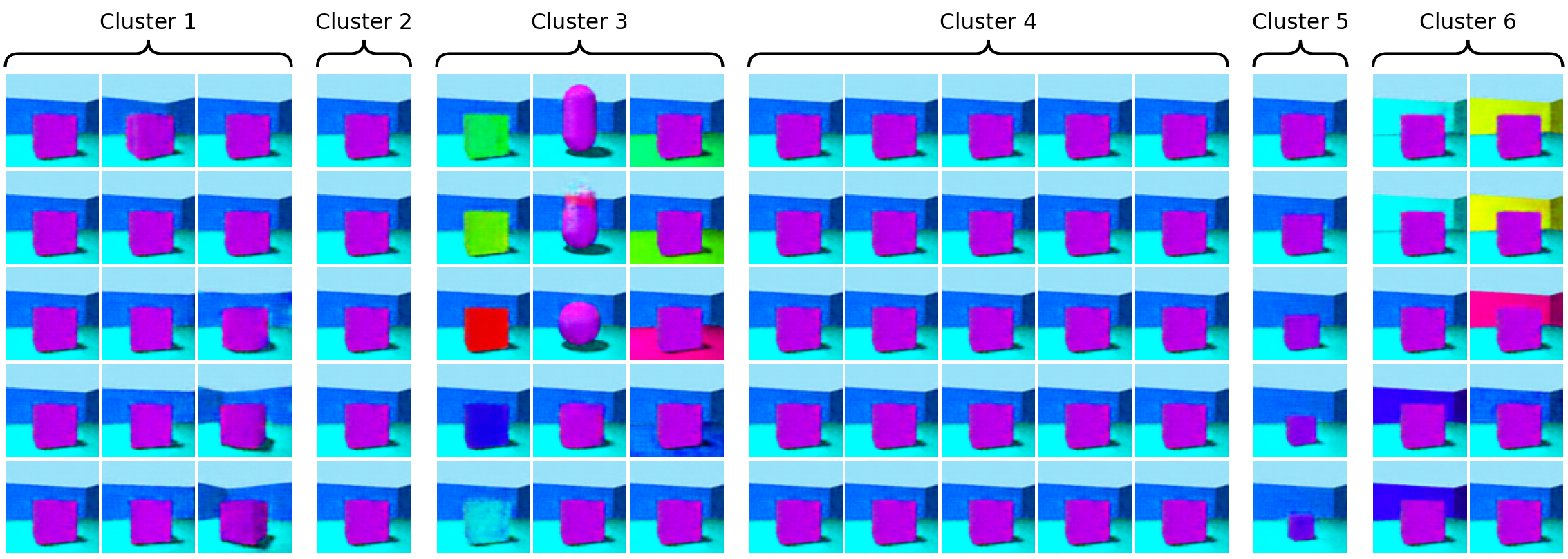}
  \caption{Latent traversals of the LBM-VAE on Shapes3D. Each column varies one
  of the 15 latent coordinates and each row one traversal value. Columns are
  grouped by learned cluster.}
  \label{fig:traversals}
\end{figure}

We conduct two experiments: clustering on MNIST~\cite{lecun2010mnist}, which
evaluates the row partition against the digit classes (ARI, NMI, accuracy), and
disentanglement on Shapes3D~\cite{3dshapes18}, which evaluates the
block-structured representation against VAE-based baselines
(MIG~\cite{NEURIPS2018_1ee3dfcd}, SAP~\cite{Kumar2017VariationalIO},
DCI~\cite{eastwood2018a}).
\\
\noindent\textbf{Implementation settings.} On MNIST, the encoder has three
stride-2 convolutional layers (32, 32 and 64 channels) followed by two
fully connected layers of 256 units, with ReLU activations and a mirrored
decoder; we use $p=20$, $K=10$ and $L=6$. On Shapes3D, all models use the
convolutional encoder and decoder of~\cite{Locatello2019}, so that differences
in scores come from the prior and the objective. Baselines use $p=10$ as
in~\cite{Locatello2019}, $\beta=6$ and, for the HFS
variants~\cite{roth2023disentanglement}, $\gamma=10$, among the values
considered in~\cite{uscidda2024disentangled}. The LBM-VAE uses $p=15$, since the
column partition becomes trivial when $p$ is close to $L$, with at most $L=6$ blocks (number of ground-truth factors) and $K=4$ row clusters (number of object shapes). Models are trained
with Adam (learning rate $5\times10^{-4}$, batch size 256) for 200 epochs, the
first 30 as a standard VAE before the LBM prior is activated, on one
NVIDIA H100 GPU of the Jean Zay supercomputer (IDRIS). Scores are reported as
mean $\pm$ std over 10 seeds at the final epoch.
\\
\noindent\textbf{Clustering results.} On the MNIST test set, the row partition
reaches an ARI of $67.4 \pm 10.3\%$, an NMI of $78.7 \pm 6.2\%$ and an accuracy
of $75.3 \pm 8.3\%$. These scores remain below those of dedicated deep clustering models such as
VaDE~\cite{3172077.3172161}, which is expected: co-clustering is a harder
problem, since the blocks are unknown and each row cluster is described by a
single Gaussian per block. Figure~\ref{fig:tsne} shows, on a separate run with only 5 pretraining epochs,
digit classes separating once the LBM prior is activated.
\noindent\textbf{Disentanglement results.} Figure~\ref{fig:traversals} shows the
learned blocks. Clusters 1 and 6 illustrate the intended behaviour: the
coordinates of Cluster 1 act only on the object orientation, and both
coordinates of Cluster 6 control the wall hue. Cluster 5 encodes the object
scale, and Cluster 4 gathers coordinates that are left unused by the decoder.
Cluster 3 still mixes object hue, shape and floor hue.

Table~\ref{tab:disentanglement} reports the scores. Without additional regularization, LBM-VAE is competitive with the best
baselines: it obtains the highest mean SAP and DCI-D and comparable
completeness and informativeness, while its MIG score remains below those of
the HFS variants.
This is
consistent with a block-wise encoding: DCI-D only requires each coordinate to
inform a single factor, which holds when a factor spans a block, whereas MIG
rewards concentrating each factor in one coordinate.

\begin{table}[!t]
\centering
\setlength{\tabcolsep}{3pt}
\resizebox{\textwidth}{!}{%
\begin{tabular}{lccccc}
\toprule
 & & & \multicolumn{3}{c}{DCI} \\
\cmidrule(lr){4-6}
Model & MIG & SAP & D & C & I \\
\midrule
$\beta$-VAE         & $0.268 \pm 0.079$ & $0.077 \pm 0.037$ & $0.603 \pm 0.129$ & $0.522 \pm 0.118$ & $0.926 \pm 0.045$ \\
$\beta$-TCVAE       & $0.312 \pm 0.171$ & \underline{$0.092 \pm 0.030$} & $0.741 \pm 0.085$ & $0.637 \pm 0.084$ & $0.977 \pm 0.015$ \\
$\beta$-VAE + HFS   & $\boldsymbol{0.616 \pm 0.181}$ & $0.085 \pm 0.055$ & $0.765 \pm 0.179$ & $\boldsymbol{0.788 \pm 0.109}$ & $0.920 \pm 0.093$ \\
$\beta$-TCVAE + HFS & \underline{$0.534 \pm 0.204$} & $0.078 \pm 0.029$ & \underline{$0.848 \pm 0.079$} & $0.779 \pm 0.088$ & $\boldsymbol{0.991 \pm 0.007}$ \\
\midrule
LBM-VAE             & $0.452 \pm 0.124$ & $\boldsymbol{0.129 \pm 0.027}$ & $\boldsymbol{0.899 \pm 0.107}$ & \underline{$0.780 \pm 0.095$} & \underline{$0.990 \pm 0.018$} \\
\bottomrule
\end{tabular}}
\caption{Disentanglement scores on Shapes3D (best mean in bold, second best underlined).}
\label{tab:disentanglement}
\end{table}

\section{Conclusion}
We introduced LBM-VAE, a variational autoencoder whose Latent Block Model prior
jointly co-clusters observations and latent dimensions, using coordinate-ascent
variational updates and a memoized update of the column partition. The learned
blocks group latent coordinates that capture common factors of variation,
yielding competitive disentanglement without any additional regularization
term. A limitation is that all latent blocks share a single row partition, whereas
distinct factors may induce distinct clusters of the observations. Future work
will therefore allow each block to learn its own observation partition, thereby extending LBM-VAE towards a multiview clustering, unlike
MFCVAE~\cite{NEURIPS2021_48cb136b} where the views are predefined by the user.
\section*{Acknowledgments}
This work was carried out within IRT VEDECOM, with the support of its industrial
partners Stellantis and Renault Group, and in collaboration with the DAVID
laboratory, UVSQ, Université Paris-Saclay. It was supported by the French
government under the "France 2030" 
operated by the French National Research Agency.
This work was granted access to the HPC resources of IDRIS under the allocation
2026-AD011017901R1 made by GENCI, on the Jean Zay supercomputer.
\bibliographystyle{unsrt}  
\bibliography{references}

\end{document}